\documentclass[11pt,a4paper]{article}
\usepackage[hyperref]{emnlp2020}
\usepackage{times}
\usepackage{latexsym}

\usepackage{microtype}
\usepackage{booktabs}
\usepackage{epsfig}
\usepackage{graphicx}
\usepackage{subcaption}
\usepackage{xspace}

\aclfinalcopy 
\def\aclpaperid{42} 

\newcommand{\eat}[1]{}

\title{Towards Accurate and Reliable Energy Measurement of NLP Models}

\author{Qingqing Cao, Aruna Balasubramanian, Niranjan Balasubramanian
\\
Department of Computer Science\\
Stony Brook University\\
Stony Brook, NY 11794, USA \\
\texttt{\{qicao,arunab,niranjan\}@cs.stonybrook.edu}
}
\date{}

\begin{document}
\maketitle

\begin{abstract}
Accurate and reliable measurement of energy consumption is critical for making well-informed design choices when choosing and training large scale NLP models. In this work, we show that existing software-based energy measurements are not accurate because they do not take into account hardware differences and how resource utilization affects energy consumption. We conduct energy measurement experiments with four different models for a question answering task. We quantify the error of existing software based energy measurements by using a hardware power meter that provides highly accurate energy measurements. Our key takeaway is the need for a more accurate energy estimation model that takes into account hardware variabilities and the non-linear relationship between resource utilization and energy consumption. We release the code and data at \url{https://github.com/csarron/sustainlp2020-energy}. 
\end{abstract}

\section{Introduction}




State-of-the-art NLP models of today ~\cite{devlin2019BERTPretraining,liu2019RoBERTaRobustly,raffel2020ExploringLimits} consume large amounts of energy. Such high-levels of energy consumption adds to the worsening global warming and can cause significant social health and safety impacts~\cite{GlobalWarminga,rolnick2019TacklingClimatea}. Recent studies have raised awareness of the carbon footprints and potential energy impacts and suggest ways to estimate and  reduce consumption~\cite{strubell2019EnergyPolicy,schwartz2019GreenAIa}. 

The success of these and future efforts depend on our ability to accurately and reliably estimate the energy consumption of NLP models. A common technique to predict the energy consumption is to measure the utilization of hardware components involved in the computation---the CPU, the GPU, and memory. 
Each of these components is associated with a single power counter value that is provided by the underlying hardware; this power counter represents the power drawn of a given component. The total energy consumption is computed as the sum of the (utilization $\times$ power counter) of the CPU, GPU, and memory, which is then adjusted by a compensation constant~\cite{henderson2020SystematicReporting,strubell2019EnergyPolicy}. We call this technique software-based power measurement.


However there are two potential sources of inaccuracies in the software-based power measurement techniques. First, the software tools are known to be inaccurate because they only consider the energy consumed by three specific hardware components, which may not reflect the energy consumption of the entire system. Second, accurately mapping hardware utilization to the energy consumption is a difficult problem. The mapping depends on the underlying hardware make and type,  energy  is not always linearly related to the utilization~\cite{pathak2011Finegrainedpower,pathak2012Whereenergya}, and  energy consumption often continues even after the NLP model has finished running~\cite{burtscher2014MeasuringGPUa}.

In this work, we use a hardware power meter  to measure ground truth energy consumption, which is more accurate.  Our goal is to quantify how far software-based measurements are from the hardware energy measurements. We compare the energy estimates obtained using prior software based models for four Transformer-based NLP models fine-tuned for a question answering (QA) task.

In the experiments, we find that (1) software energy estimates can differ from the hardware power measurements by 20\% on average. Further, the standard deviations are 2$\times$ larger than hardware power meters. (2) Power-models need to take into account the underlying hardware, make, and configuration. Hardware-agnostic energy measurements results in large errors, for example, when applied to machines with different configurations (e.g. different GPU models, \# of GPUs used).

Finally, we show the importance of accurate power-models to make the right accuracy/energy trade-off. Ground-truth energy measurements using a hardware meter show that RoBERTa-base incurs 13\% more energy on average. But RoBERTa-base can answer 2.2\% more questions correctly over BERT-base. However, existing power-models estimate the additional power consumption of RoBERTa-base to be 25\%. Such inaccuracies can lead to wrong  conclusions and poor optimizations for model practitioners. The results in this paper suggests that we need better estimation models that are calibrated to account for hardware variabilities and the non-linear relationship between power consumption and resource utilization.

\section{Experiments Methodology}
\label{sec:methodology}

In this section, we describe our setup and methodology for energy measurements.  We focus on energy consumption of inference for a QA task using a hardware power meter. For comparison purposes, we track software reported energy values as well.

\subsection{Setup}

{\noindent \bf Devices:} We use 2 GPU-equipped desktop PCs as the target hardware for running our models. See Table \ref{tab:setup} for details. 

We fine-tune and perform inference in all 4 models on the SQuAD v1.1 question answering dataset~\cite{rajpurkar2016SQuAD100} using PyTorch~\cite{paszke2019PyTorchImperative} v1.6 through the HuggingFace Transformers~\cite{wolf2020HuggingFaceTransformers} library. The four models we study are --- BERT-base~\cite{devlin2019BERTPretraining}, RoBERTa-base~\cite{liu2019RoBERTaRobustly}, MobileBERT~\cite{sun2020MobileBERTCompact}, and DistillBERT~\cite{sanh2020DistilBERTdistilled}.

\begin{table}[ht]
  \setlength\tabcolsep{2pt}
  \begin{tabular}{@{}lll@{}}
  \toprule
  Specification & \textbf{PC1}     & \textbf{PC2}   \\ \midrule
  CPU           & Intel i9-7900X   & Intel i7-6800K \\
  Memory        & 32 GiB           & 32 GiB         \\
  GPU           & 2$\times$  GTX 1080 Ti & 2$\times$ GTX 1070   \\
  GPU Memory    & 11.2 GiB per GPU & 8 GiB per GPU  \\
  Storage       & 1 TiB SSD        & 1 TiB SSD      \\ \bottomrule
  \end{tabular}
  \caption{Target hardware specifications. }
  \label{tab:setup}
  \end{table}

\begin{figure*}[ht!]
  \centering
		\includegraphics[width=0.7\linewidth]{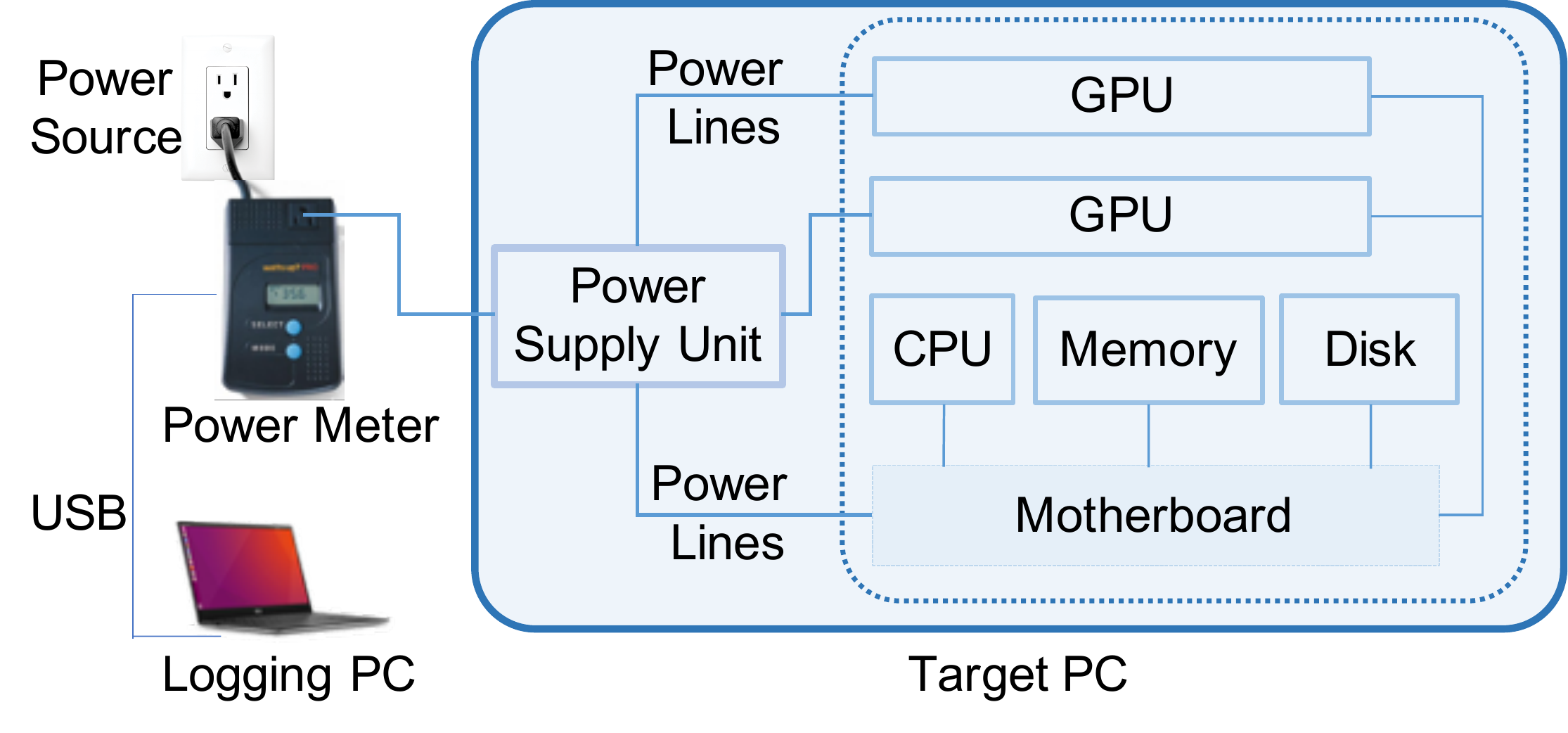}
		\caption{Illustration of the energy measurement setup using a hardware power meter. }
		\label{fig:setup}
\end{figure*}

{\noindent \bf Hardware-based Measurements} We use the WattsUP power meter~\cite{WattsUpMeter}\footnote{The device is available on Amazon \url{https://amzn.to/2EoP0tU}} to measure \emph{all} of energy consumed by a PC. The WattsUpMeter is used to power the computer, and the power meter records the passthrough current and voltage values every 1 second. This allows us to accurately measure the power draw at a 1 second granularity. Figure~\ref{fig:setup} shows the energy measurement setup.  We obtain current, voltage, and timestamp values from the power meter's built-in USB port. The energy ($e$) consumed during a time period is then calculated using the sampled current $(I_t)$ and voltage $(V_t)$ values in that period: $e = \sum_{t} V_t I_t$.

{\noindent \bf Software-based Measurements:} For comparisons, we use the software-based energy measurements provided by the {\em experiment-impact-tracker} framework~\cite{henderson2020SystematicReporting} which estimates energy as a function of the GPU, CPU, and memory utilization. More details about the model can be found in ~\S\ref{sec:estimation-modeling}.


\subsection{Methodology}

For each NLP model, we obtain the energy measurements over a random sample of 1000 questions from the SQuAD 1.1 dev split. We repeat these measurements over 10 runs and report the average and standard deviation of energy values.
We use 1 GPU to run all experiments, but show the energy measurements accuracy for multiple GPUs in \S\ref{sec:estimation-modeling}. Since it is common to batch process inputs on GPUs, we benchmark batch size 1 and batch sizes from 2 to 16 with step 2 \footnote{We tried larger batch sizes, but found the energy and latency values to be similar for batch sizes between 18 and 32, therefore, we omit numbers with batch size larger than 16 for brevity.}.

To guarantee the consistency and reliability of the hardware energy measurement, we cool down the PCs after each experiment finishes to avoid potential overheating issue that can cause subsequent energy distortions. We  measure the standby power consumption (when the CPU load is $<0.1$\%) and ensure before running the experiments that the PC does not draw more than the standby power. Further, no other application is running during our experiments. 

We record the start and end timestamp of the benchmarked program, and extract the energy values by comparing and aligning the timestamps from the power meter logs. All the energy and latency numbers are end to end, except in \S\ref{sec:interaction} where we extract the numbers for the prediction part only. In \S\ref{sec:interaction}, we study the latency speedups for model prediction, whereas the latency numbers for data loading, startup and cleanup are often comparable to model inference given that we only run for 1000 questions. In the real case, the startup and cleanup costs will be amortized if running millions or billions of the model inference.

\section{Energy Results of NLP Models}

In this section, we discuss the energy results of the four Transformer-based NLP models on a question answering task.

\subsection{Existing Software-Based Energy Measurements Are Not Accurate}


We use the energy values recorded by the hardware power meter as ground truth, and report both error percentage ($|$true energy - software-based energy measurement$|$/true energy) and standard deviations of the software energy measurements for all four NLP models. 

\begin{figure*}[ht]
	\centering{
	\begin{subfigure}{0.465\textwidth}
		\includegraphics[width=\columnwidth]{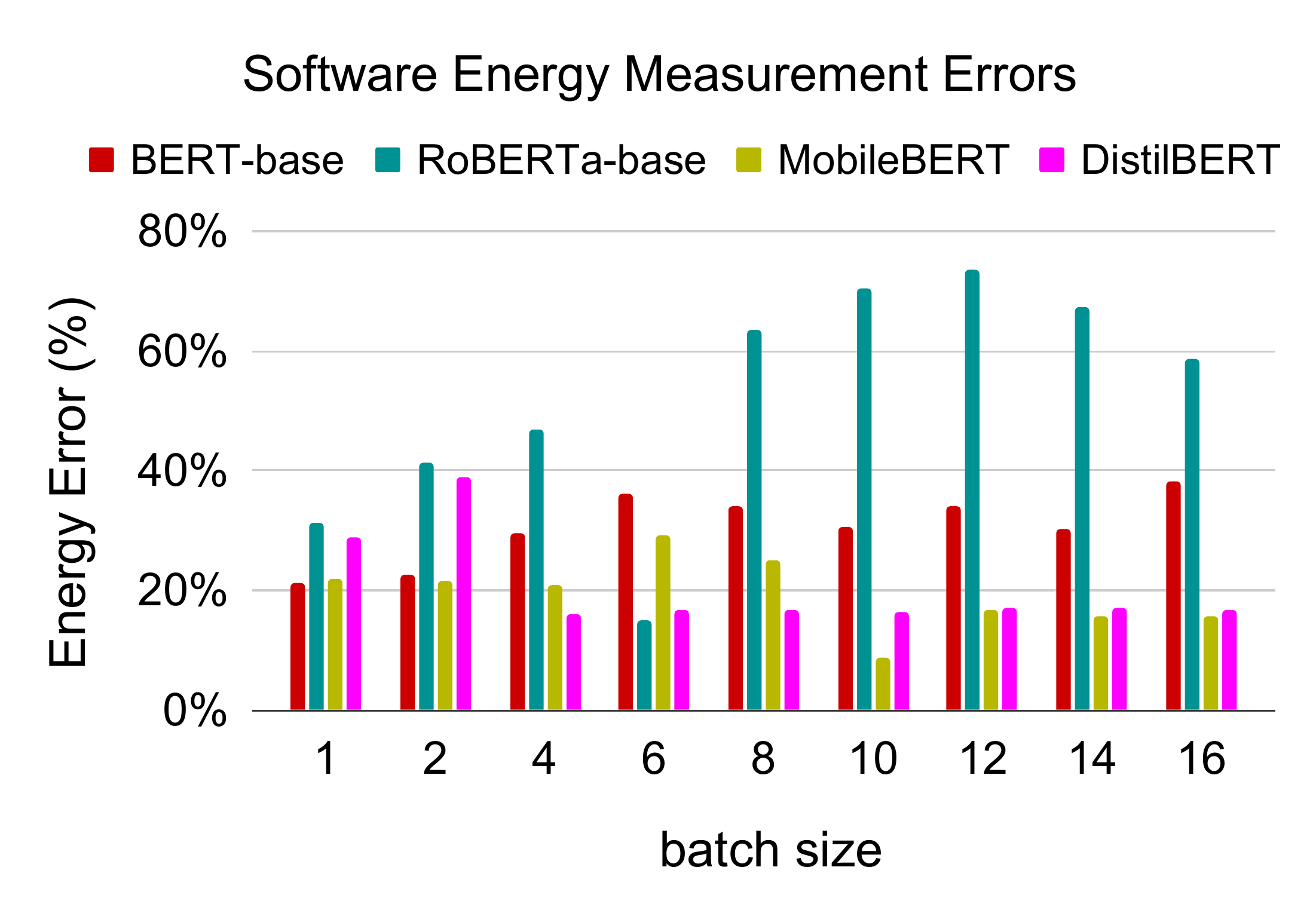}
		\caption{Errors when using software-based energy measurements for the 4 studied models. We use the hardware power meter as the ground truth energy, and compute the error as the energy differences percentage of the ground truths.}
		\label{fig:energy-error}
	\end{subfigure}
	\quad
	\begin{subfigure}{0.465\textwidth}
		\includegraphics[width=\columnwidth]{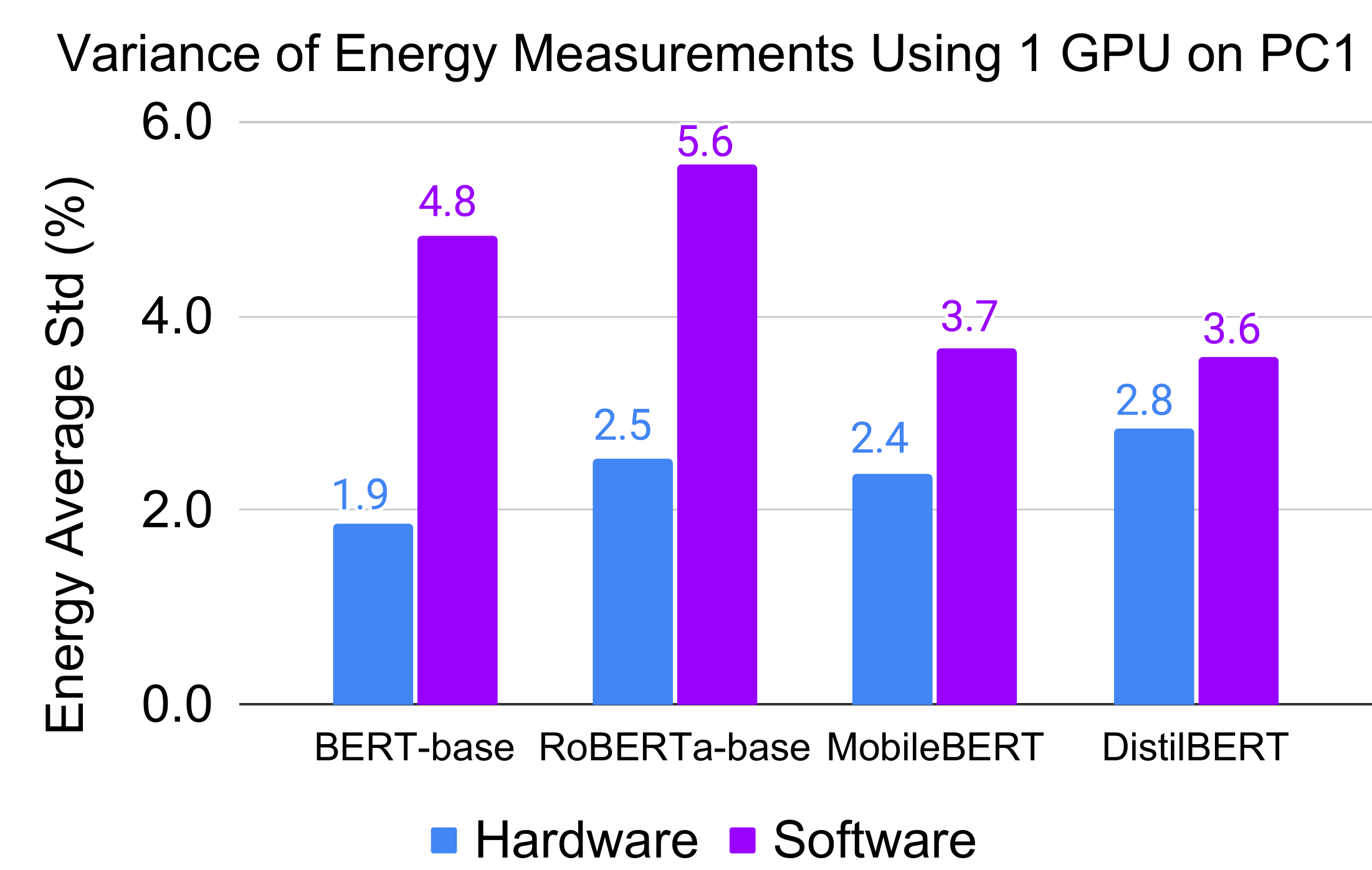}
		\caption{Standard deviations of both energy measured using hardware power meter and using software energy estimates. We compute the standard deviation across 10 runs for the 4 studied models.}
		\label{fig:energy-std}
	\end{subfigure}
	\caption{Accuracy and robustness comparison between hardware and software based energy measurements. We use 1 GPU on PC1 for all the experiments.}
	\label{fig:pc1-energy}
	}
\end{figure*}

Figure \ref{fig:energy-error} shows that the error of the software measurements ranges from 2\% to as much as 47\%. In more than 90\% of the runs the error is at least 20\%, and for a fifth of the runs the error is at least 30\%. On average the error percentages are substantial for all models --- error on BERT-base is 26\%, RoBERTa-base is 47\%, MobileBERT is 30\%, and DistilBERT is 36\%. While there are some settings where software measurements is accurate (for example, the error is only 2.7\% for RoBERTa-base model with batch size 2), it is not accurate in general.



Figure \ref{fig:energy-std} shows that the standard deviations for software energy measurements are twice as large as that of hardware-based energy measurements. Large deviations for different runs of a model in the same setting makes the measurements unreliable. 


The main takeaway here that existing software-based energy measurements can be substantially inaccurate. However, they are more convenient to estimate energy consumption compared to using hardware power meters. Going forward, we need to design more accurate software measurements that come close to the ground truth. 

\subsection{Energy Measurements Using Hardware Agnostic Parameters Is Suboptimal}
\label{sec:estimation-modeling}

Why are existing software-based energy measurements~\cite{strubell2019EnergyPolicy,henderson2020SystematicReporting} not accurate? 
The software-based energy model computes energy by aggregating resource usage as follows: $e_{total} = PUE \sum_p (p_{dram}e_{dram} + p_{cpu}e_{cpu} + p_{gpu}e_{gpu})$, where $p_{resource}$ \footnote{$resources$ can be $dram$, $cpu$, $gpu$} are the percentages of each system resource used by the attributable processes relative to the total in-use resources and $e_{resource}$ is the energy usage of that resource. The constant for power usage effectiveness (PUE) compensates for extra energy used to cool or heat data centers. 

There are two potential problems in this linear energy model. First, different hardware devices (e.g. different CPU or GPU models, different number of GPUs connected, etc.) can have different cooling or heating effects causing large variations in the amounts of energy consumed. However, the energy model uses the PUE constant as a hardware agnostic parameter, which does not account for such differences in device specifications. This makes the final energy measurements less reliable. Second, assigning energy credits based on process resources is not always reliable. CPUs and GPUs often have power lags, power distortions, and tail energy especially during starting new processes or finishing existing processes~\cite{burtscher2014MeasuringGPUa,krzywda2018Powerperformancetradeoffs}. 

We conducted two empirical experiments to study these problems: (1) measuring the energy consumption of running the 4 NLP models on two different machines -- {\bf PC1} and {\bf PC2}. The detailed device information is described in \S\ref{sec:methodology}. (2) Use two GPUs on {\bf PC1} to perform inference for the 4 NLP models instead of a single GPU. 

\begin{figure}[ht!]
  \centering
		\includegraphics[width=\columnwidth]{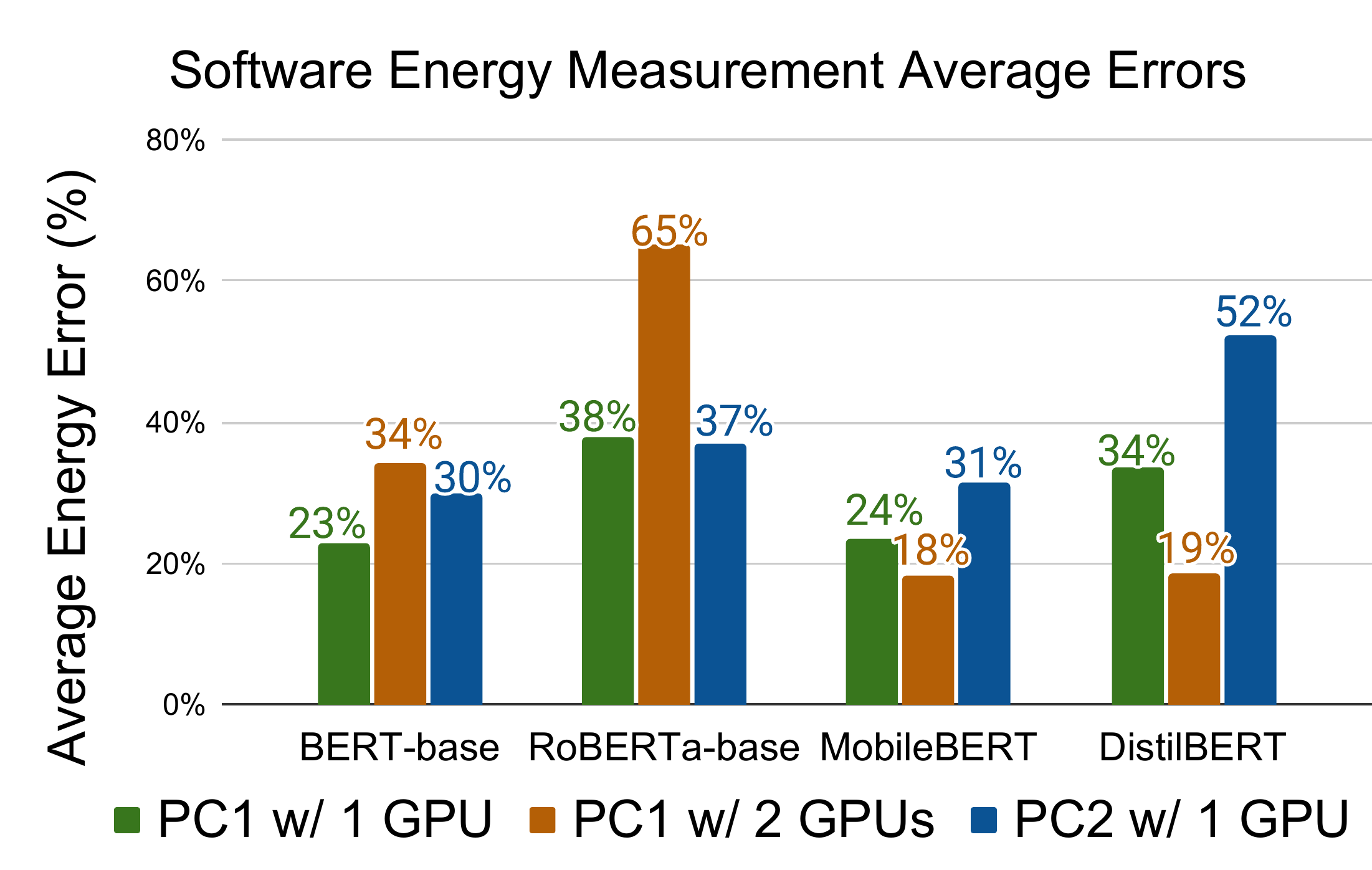}
		\caption{Average energy error of the software measurements for the 4 studied models using different hardware device configurations. The error patterns are different across all 3 settings, for example, (1) using two GPUs (instead of one) on the same machine can cause more errors; (2) using the same number of GPUs but with different hardware specifications may lead to different energy errors. (i.e., compare using 1 GPU on PC2 to 1 GPU on PC1)}
		\label{fig:diff-energy-error}
\end{figure}

\begin{figure}[ht!]
	\centering{
	\begin{subfigure}{0.465\textwidth}
		\includegraphics[width=\columnwidth]{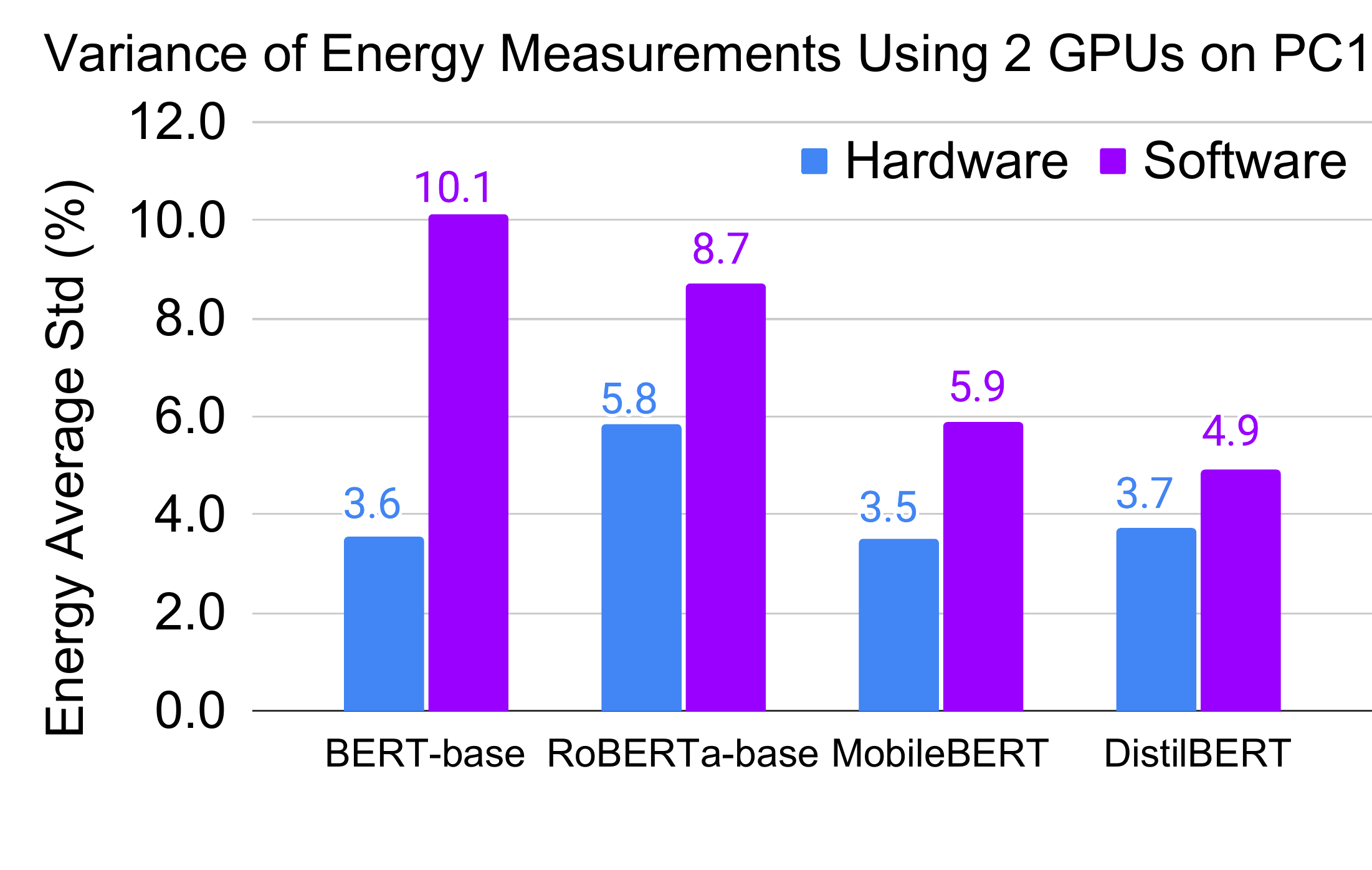}
		\caption{Standard deviations of the hardware power meter measurements and software energy measurements using 2 GPUs on PC1 to perform inference.}
		\label{fig:2gpu-energy-std}
	\end{subfigure}
	\quad
	\begin{subfigure}{0.465\textwidth}
		\includegraphics[width=\columnwidth]{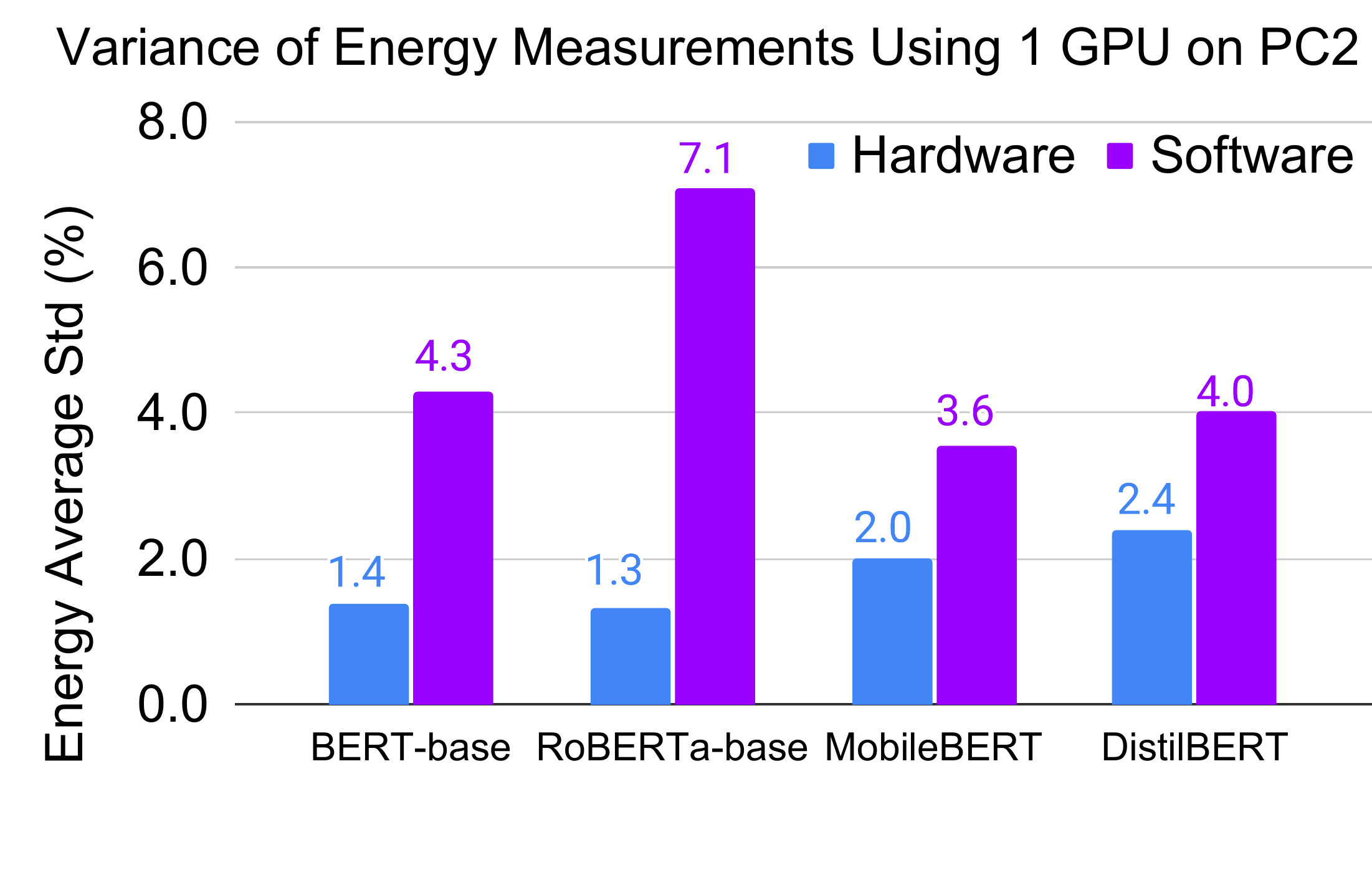}
		\caption{Standard deviations of the  hardware power meter measurement and software energy measurements 1 GPU on PC2 to perform inference.}
		\label{fig:pc2-energy-std}
	\end{subfigure}
	\caption{Comparing the standard deviation in energy estimates under different hardware device configurations. }
	\label{fig:different-hw-energy}
	}
\end{figure}

Figure \ref{fig:diff-energy-error} shows that the energy errors are prominent when using two GPUs for inference compare to one-GPU setting or using a different GPU model. This is likely because the linear energy estimate model cannot easily take into account the above energy factors (power lag, distortion and tail energy) that affect GPU resources usage. 
Variable PUE can possibly address this, but that requires careful calibration based on the ground truth energy from the hardware power meters. 
Figure \ref{fig:different-hw-energy} shows the standard deviation when using existing software-based energy measurements.

\subsection{Software-Based Energy Measurements Can Lead to Bad Design Choices}
\begin{figure}[ht]
	\centering{
	\begin{subfigure}{0.465\textwidth}
		\includegraphics[width=\columnwidth]{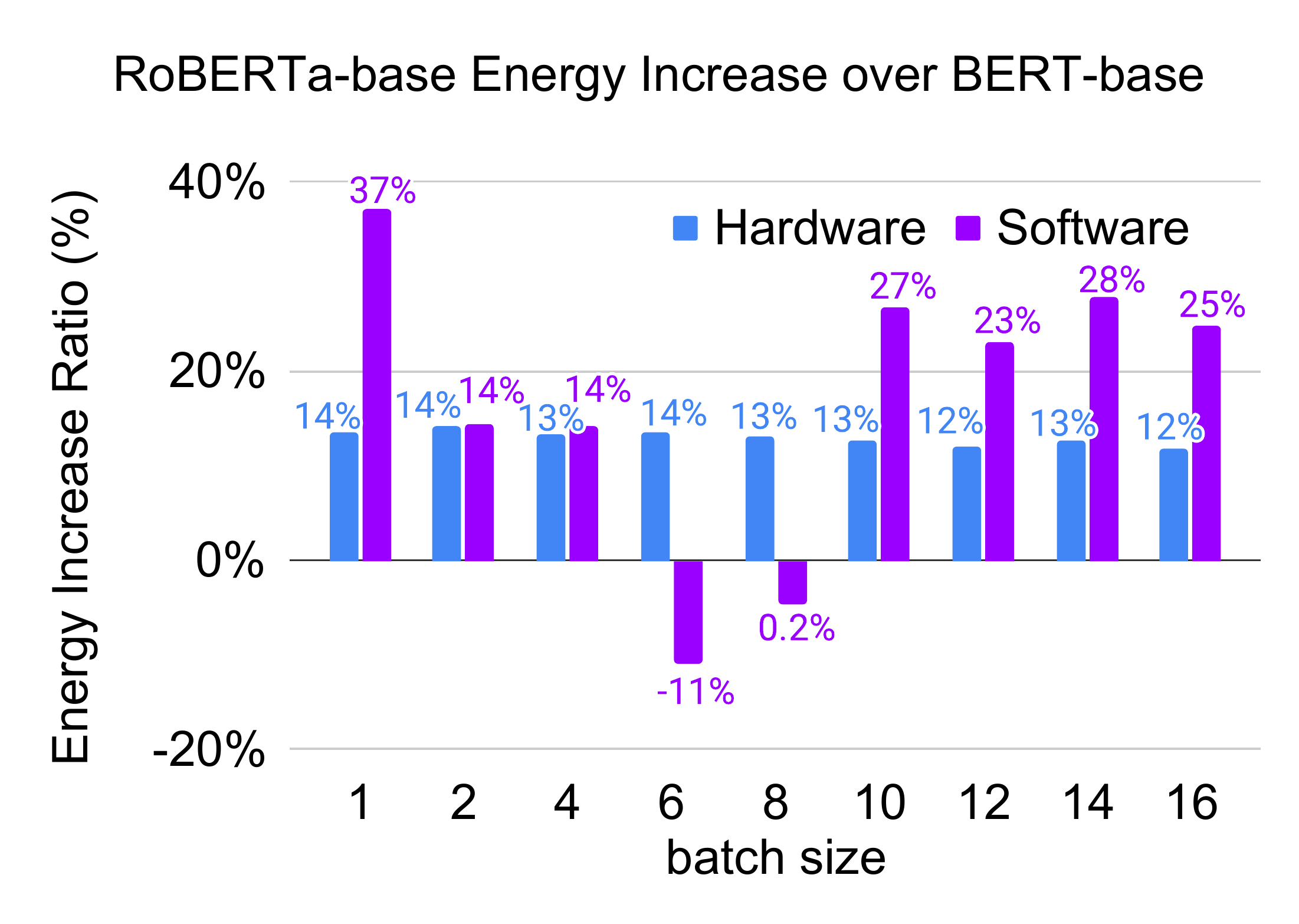}
		\caption{Energy increase ratios from DistilBERT to MobileBERT.}
		\label{fig:energy-increase-ratio-hw}
	\end{subfigure}
	\quad
	\begin{subfigure}{0.465\textwidth}
		\includegraphics[width=\columnwidth]{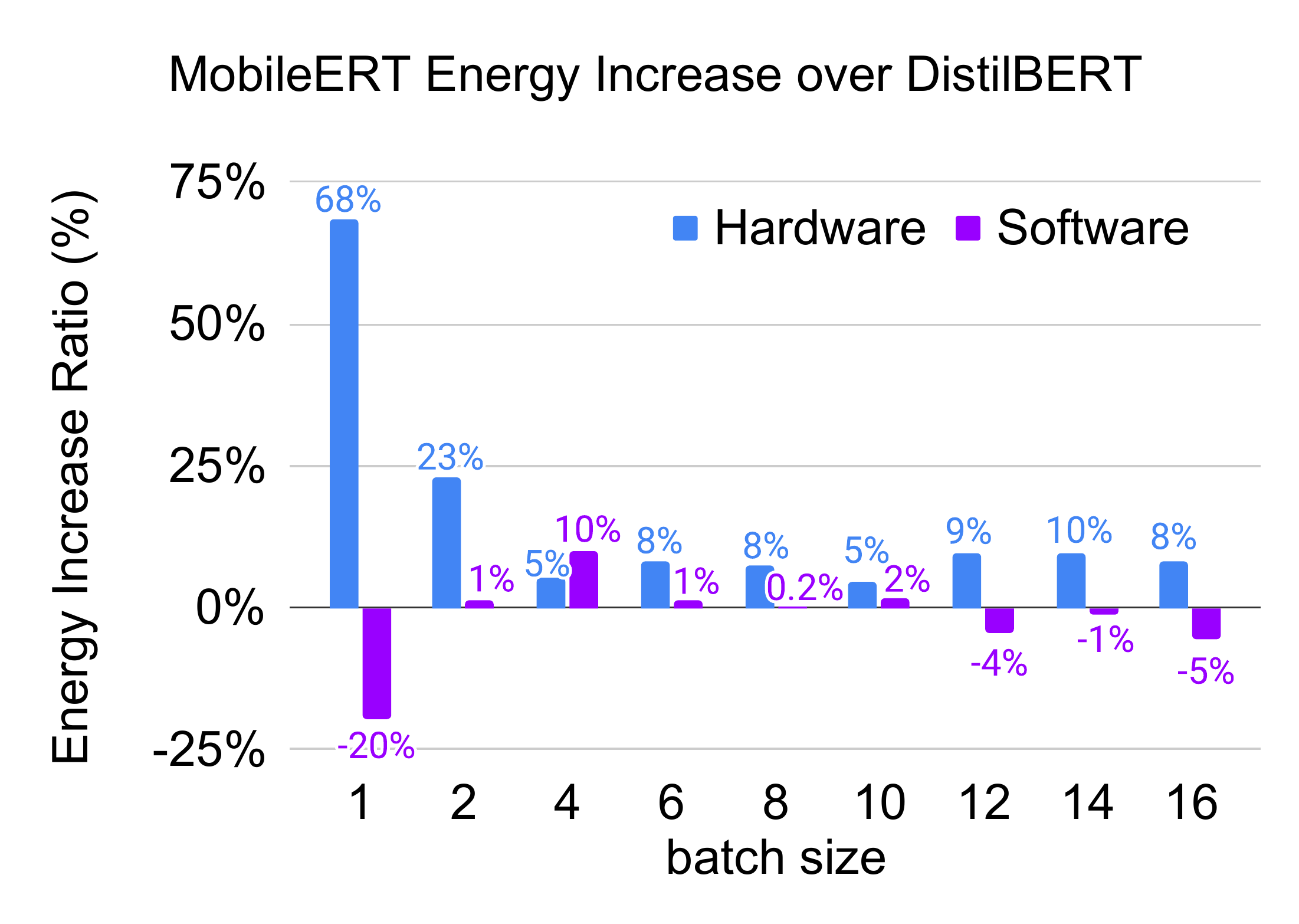}
		\caption{Energy increase ratios from BERT-base to RoBERTa-base.}
		\label{fig:energy-increase-ratio-sw}
	\end{subfigure}
	\caption{Energy increase ratio comparison using hardware and software measurements.}
	\label{fig:energy-increase-ratio}
	}
\end{figure}

The inaccuracy and robustness issues in software-based measurements can adversely impact model choices when considering energy and effectiveness trade-offs. To demonstrate this we consider two decision problems. One where we want to choose between BERT-base with RoBERTa-base, and another problem where we want to choose between MobileBERT and DistilBERT. Table \ref{tab:score} summarizes the performance scores of these models on the SQuAD 1.1 QA dataset. Figure~\ref{fig:energy-increase-ratio-hw} shows that RoBERTa-base correctly answers an additional 2.2\% questions over BERT-base but it incurs 13\% more energy on average. Similarly, MobileBERT answers 3.5\% more questions correctly with 13\% more energy budget compared to DistilBERT. Moreover, for MobileBERT and DistilBERT, batching questions help close the relative gap of energy costs.

If we instead use software-based energy measurements, however, presents a misleading picture. According to software energy measurements shown in Figure \ref{fig:energy-increase-ratio-sw}, RoBERTa even consumes less energy than BERT (batch sizes 6 and 8), and MobileBERT can be more energy efficient than DistillBERT for many batch sizes (1, 12, 14, 16). Neither conclusion is true.


\begin{table}[ht]
	\centering
	\setlength\tabcolsep{10pt}
	\begin{tabular}{@{}lll@{}}
		\toprule
		Model        & EM   & F1-score \\ \midrule
		BERT-base    & 80.8 & 88.2     \\
		RoBERTa-base & 83.0 & 90.4     \\ \midrule
		MobileBERT   & 82.6 & 90.0       \\
		DistilBERT   & 79.1 & 86.8     \\ \bottomrule
	\end{tabular}
	\caption{SQuAD 1.1 task performance scores of the 4 studied models.}
	\label{tab:score}
\end{table}

\subsection{Interactions between Inference Latency and Energy Consumption Are Non-trivial}
\label{sec:interaction}
With the more accurate hardware energy measurements, we investigate the relationship between latency and energy consumption. 
In particular, we correlate the model energy consumption with its inference latency and task-specific performance. 
Note that, in this section, to better characterize the model inference latency and energy interactions, we do not use the end to end latency and energy numbers. Instead, we focus on the model prediction process, i.e. right before the model runs prediction and after the model finishes the prediction of all examples. 

\begin{figure}[ht!]
  \centering
		\includegraphics[width=\columnwidth]{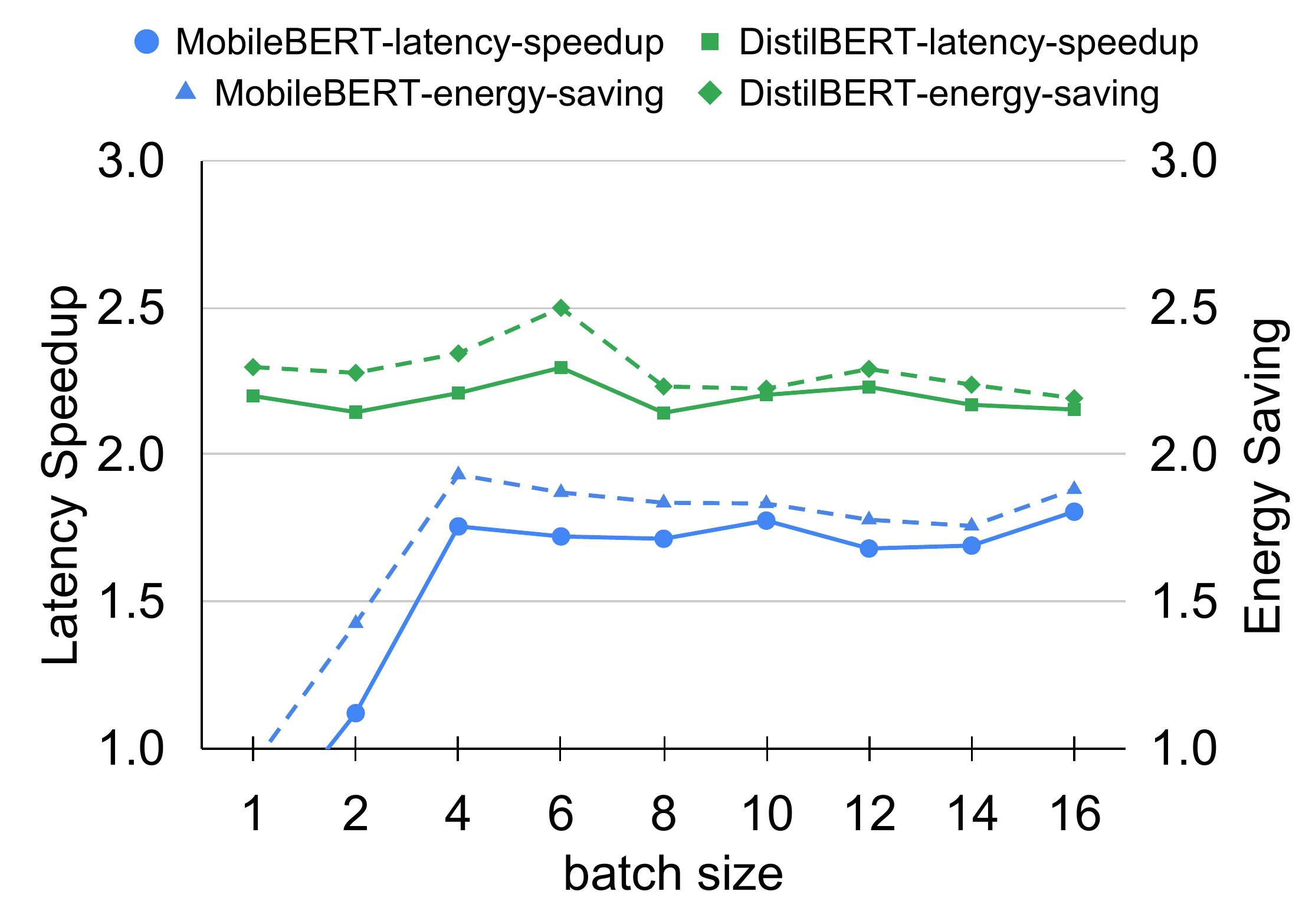}
		\caption{Latency speedup versus energy savings. All numbers are relative to the RoBERTa-base model. We report the hardware based energy values. Since the \textit{experiment-impact-tracker} software does not sample sufficient energy values, we cannot extract the software energy for the prediction process only, hence omit comparison. }
		\label{fig:latency-energy}
\end{figure}

Figure \ref{fig:latency-energy} shows the inference latency speedup versus energy savings of MobileBERT and DistilBERT models over the RoBERTa-base model. We can see that smaller batch sizes ($<10$) give more energy benefits compared to latency improvement, but as the inference batch size increases, the latency and energy savings are approximately proportional. This is beneficial to mobile settings where smaller batch sizes happen more frequently (e.g., users ask a question at a time instead of asking many questions simultaneously). 

\section{Related Work and Discussion}
\label{sec:related}

Energy estimation is an important research topic in both the machine learning and system community. 
We discuss two threads of research related to energy estimation for NLP models:

{\bf Energy estimation in machine learning and NLP.} \citet{henderson2020SystematicReporting} use a software framework called \textit{experiment-impact-tracker} to report the aggregated energy of benchmark programs. The \textit{experiment-impact-tracker} collects hardware resources statistics, and uses a simple linear model~\cite{strubell2019EnergyPolicy} to estimate the total energy where the coefficients are fixed to a constant without considering the actual hardware device configurations. We have shown in the experiments that such software based energy estimation methods are neither accurate nor robust. We recommend using hardware power meters to measure the energy consumption, and then possibly calibrate the software energy values. \cite{zhou2020HULKEnergy} presents an energy efficient benchmark for NLP models. However, they only report the time (hours) and cost (dollars) for training and testing NLP models, the actual energy numbers remain unknown. The Green AI \cite{schwartz2019GreenAIa} work suggests using metrics like floating point operations (FPO) to measure energy efficiency. However, \citet{henderson2020SystematicReporting} argues such metrics alone cannot accurately reflect energy consumption. \citet{garcia-martin2019Estimationenergy} provide a comprehensive survey of energy estimation methods in machine learning, but no energy measurements for NLP models were reported. 

{\bf Energy modeling for systems and applications.}
Energy estimation for battery powered devices such as mobile phones is critical since mobile applications utility can be limited by the battery life. Previous work \cite{pathak2011Finegrainedpower,pathak2012Whereenergya,yoon2012AppScopeapplication,cao2017DeconstructingEnergy} study various fine-grained system-level power modeling and profiling techniques to help understand energy drain of applications. NLP models essentially power many emerging applications such as personal assistants with mobile intelligence. However, the energy implications for these NLP models are not studied. It is unclear how to apply the existing energy estimation methods for mobile applications to NLP models. We believe it is important to understand the computational semantics in NLP models before leveraging these existing power modeling methods. Using fine-grained power estimation models and profiling techniques could further improve our understanding of how NLP models consume energy and is an interesting future work.

{\bf Limitations.} In this work, we collect the energy values every 1 second, which can reflect the total amount of energy consumed for batched inferences that often last over 10 seconds. However, if one needs to understand the energy spent inside the model for a single inference, the energy values are still coarse-grained, we will explore fine-grained energy measurement solutions to study this issue in the near future. More complex energy issues like tail power, energy distortion~\cite{burtscher2014MeasuringGPUa,pathak2011Finegrainedpower} also affect hardware power meters if analyzing the energy spent inside the model. Further, it is not clear yet which machine component (CPU, GPU or memory reads/writes) takes how much energy for the NLP models. We leave this to future work.

\section{Conclusions}
As NLP models keep getting larger, reducing the energy impact of deploying these models is critical. Recent work has enabled estimating and tracking the energy of these NLP models. These works design a software-based technique to estimate energy consumption by tracking resource utilization. However, we show that currently used software-based measurements method is not accurate. We use a hardware power meter to accurately measure energy and find that this measurement method has an average error of 20\% and can lead to making inaccurate design choices. Going forward, we hope this paper encourages the NLP community to build on current systems research to design more accurate energy models that take into account the underlying power dynamics and device variabilities. 





\bibliographystyle{acl_natbib}
\bibliography{ref}


\end{document}